\documentclass[conference]{IEEEtran}
\usepackage{graphicx}
\usepackage{amsmath}
\usepackage{amssymb}
\newcommand{\C}{\mathcal{C}}
\begin{document}

\title{Navigation Through Endoluminal Channels Using Q-Learning}

\author{\IEEEauthorblockN{Oded Medina}
\IEEEauthorblockA{\textit{Faculty of Engineering} \\
\textit{Ariel University}\\
Ariel, Israel \\
OdedMedina@gmail.com}
\and
\IEEEauthorblockN{Liora Kleinburd}
\IEEEauthorblockA{\textit{Faculty of Engineering} \\
\textit{Ariel University}\\
Ariel, Israel \\
LioraBurd@gmail.com}
\and
\IEEEauthorblockN{Nir Shvalb}
\IEEEauthorblockA{\textit{Faculty of Engineering} \\
\textit{Ariel University}\\
Ariel, Israel \\
NirSh@ariel.ac.il}
}

\maketitle

\begin{abstract}
In this paper, we present a novel approach to navigating endoluminal channels, specifically within the bronchial tubes, using Q-learning, a reinforcement learning algorithm. The proposed method involves training a Q-learning agent to navigate a simulated environment resembling bronchial tubes, with the ultimate goal of enabling the navigation of real bronchial tubes. We discuss the formulation of the problem, the simulation environment, the Q-learning algorithm, and the results of our experiments. Our results demonstrate the agent's ability to learn effective navigation strategies and reach predetermined goals within the simulated environment. This research contributes to the development of autonomous robotic systems for medical applications, particularly in challenging anatomical environments.
\end{abstract}

\section{Introduction}
Endoluminal navigation, particularly in anatomical channels such as the bronchial tubes, presents a unique set of challenges in the field of medical robotics. The intricate and dynamic nature of these channels requires precise and adaptable navigation techniques to ensure safe and effective procedures. Traditional manual navigation methods can be time-consuming and challenging, often necessitating highly skilled operators. Therefore, there is a growing interest in developing autonomous robotic systems that can navigate endoluminal channels with minimal human intervention.

The goal of this research is to investigate the feasibility of using Q-learning, a well-established reinforcement learning algorithm, to enable autonomous navigation within endoluminal channels. Q-learning has shown success in various domains, such as game-playing and control tasks, making it a promising candidate for the complex task of navigating within anatomical structures. The use of Q-learning in medical robotics, particularly for endoluminal navigation, is relatively unexplored, and this paper aims to contribute to filling this gap.

\subsection{Problem Statement}
Navigating through endoluminal channels, such as the bronchial tubes, requires addressing several key challenges. The channels are narrow, curved, and may contain branching pathways, making it difficult to maintain the correct trajectory. Additionally, depth perception can be limited, further complicating navigation. Autonomous navigation systems must account for real-time obstacles, potential collisions with channel walls, and the need to reach specific target locations. Notably, it's challenging to achieve precise navigation solely through local maneuvers. More often, it's evident that in order to turn or change direction, actions must be initiated well before reaching a bifurcation. This foresight is necessary to properly lean on the cavity walls and manage to take the turn effectively.

\subsection{Motivation and Objectives}
The motivation for this research stems from the potential benefits of autonomous robotic navigation within endoluminal channels. By developing intelligent navigation algorithms, we can enhance the precision, efficiency, and safety of medical procedures performed within these channels. This is particularly relevant for applications such as bronchoscopy, where accurate navigation is critical for diagnosing and treating respiratory conditions.

The main objectives of this research are as follows:
\begin{itemize}
\item Develop a simulation environment that closely mimics the conditions of endoluminal channels, specifically bronchial tubes.
\item Implement a Q-learning algorithm to train an agent to autonomously navigate within the simulated environment.
\item Evaluate the agent's performance in terms of successful navigation and the strategies it learns.
\item Discuss the implications and potential applications of the proposed approach in the context of medical robotics.
\end{itemize}

\subsection{Related Work}
Motion planning is a crucial field within robotics. For problems involving high-dimensional scenarios, such as flexible wires, it's often necessary to address the issue within the configuration space \cite{medina2013c}. The chosen actions might be contingent upon the current configuration, and in instances where the configuration space is fully understood, it can be resolved analytically \cite{shvalb2007motion}. In the context of non-Markovian challenges, one might resort to heuristics, leveraging intuitive approaches and shortcuts to hasten the path-finding procedure \cite{masuri2020gait}. In this study, we focus on a high-dimensional non-Markovian problem where the configuration space is unknown.

Previous research in medical robotics and autonomous navigation has primarily focused on various imaging modalities, sensor integration, and path planning algorithms. Bronchoscopy, a commonly used medical procedure involving the insertion of a bronchoscope into the airways, has seen advancements in imaging techniques and robotics-assisted navigation \cite{smith2019advances}\cite{burgner2015continuum}. Imaging modalities such as fluoroscopy and monocular cameras have been explored to aid navigation and provide real-time feedback \cite{webster2006toward}\cite{li2020curvature}. However, the development of autonomous navigation algorithms using reinforcement learning, specifically Q-learning, for bronchoscopy remains relatively unexplored.

In the field of robotics, Q-learning has been applied to path planning, robotic control, and game-playing scenarios \cite{mnih2015human}\cite{sutton2018reinforcement}. These applications showcase the adaptability and learning capabilities of Q-learning algorithms. However, the translation of Q-learning to the context of navigating complex and dynamic anatomical structures presents unique challenges and opportunities.

\subsection{Contribution}
For the end described above we shall devise the following:
\begin{itemize}
\item Formulation of the endoluminal navigation problem within bronchial tubes as a reinforcement learning task.
\item Development of a simulated environment that emulates the challenges of navigating within anatomical channels.
\item Implementation and evaluation of a Q-learning agent trained to navigate within the simulated environment.
\item Discussion of the agent's learned navigation strategies and their potential implications in medical robotics.
\end{itemize}

The remainder of this paper is organized as follows: Section 2 provides an overview of reinforcement learning and Q-learning. Section 3 describes the simulation environment and the Q-learning algorithm implementation. Section 4 presents the experimental results and analysis. Section 5 discusses the significance of the results and suggests future research directions. Finally, Section 6 concludes the paper.

\section{Reinforcement Learning and Q-Learning}
\emph{Reinforcement Learning} (RL) is a machine learning paradigm concerned with learning how an agent should take actions in an environment to maximize cumulative rewards. In RL, an agent interacts with an environment by taking actions and receiving feedback in the form of rewards. The agent's goal is to learn a policy that maps states to actions in a way that maximizes the expected sum of rewards over time [7].

At the heart of RL lies the Markov Decision Process (MDP), which provides a mathematical framework for modeling decision-making problems. An MDP is defined by a tuple $(S, A, P, R)$, where:
\begin{itemize}
\item $S$ is the set of possible states $s$ in the environment.
\item $A$ is the set of possible actions the agent can take.
\item $P$ is the state transition probability function, $P(s'\vert s, a)$, which gives the probability of transitioning to state $s'$ from state $s$ after taking action $a$.
\item $R$ is the reward function, $R(s, a, s')$, which gives the immediate reward received after transitioning to state $s'$ from state $s$ by taking action $a$.
\end{itemize}

\emph{Q-learning} is a model-free reinforcement learning algorithm that aims to learn the optimal action-selection policy for an agent. The Q-learning algorithm iteratively updates estimates of the Q-values, which represent the expected cumulative reward for taking a specific action in a particular state.

The Q-values are updated using the \emph{Bellman equation}, which expresses the Q-value of a state-action pair in terms of the immediate reward and the maximum Q-value of the next state:

\[
Q(s, a) = R(s, a) + \gamma \cdot \max_{a'} Q(s', a')
\]

where \(s\) is the current state, \(a\) is the chosen action, \(s'\) is the next state, \(\gamma\) is the discount factor that balances immediate and future rewards, and \(a'\) iterates over all possible actions in the next state.

The Q-learning algorithm iteratively updates the Q-values using the following equation:

\[
Q(s, a)' \leftarrow Q(s, a) + \alpha \cdot [R(s, a) + \gamma \cdot \max_{a'} Q(s', a') - Q(s, a)]
\]

where \(\alpha\) is the learning rate, which determines the weight given to the new information compared to the existing Q-value.

A challenge in reinforcement learning is balancing exploration and exploitation. Exploration involves trying new actions to gather information about the environment, while exploitation involves choosing the best-known actions based on the learned Q-values. To address this challenge, the epsilon-greedy algorithm is commonly used.
The epsilon-greedy algorithm introduces an exploration factor \(\epsilon\) that determines the probability of taking a random action versus the best-known action. At each step, a random number is generated. If the random number is less than \(\epsilon\), the agent takes a random action; otherwise, it takes the action with the highest Q-value.

 \begin{figure}[ht]
\begin{center}
\includegraphics[width=1\columnwidth]{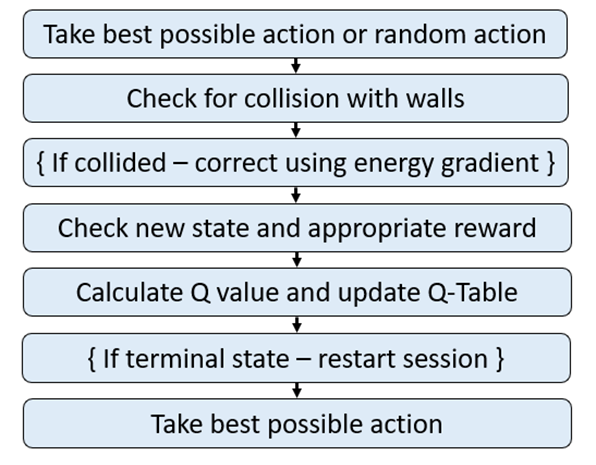}
\caption{Applying Q-learning to endoluminal autonomous navigation paradigm. }
\label{fig:1}
\end{center}
\end{figure}

In the context of endoluminal navigation, the state space is taken as the robot's position and orientation within the anatomical channel, and the action space comprises the robot's possible movements (e.g., bending, advancing). The Q-values indicate the expected cumulative reward for taking specific actions in different states, guiding the agent's navigation decisions.

\section{The mathematics model}
To investigate the feasibility of using Q-learning for endoluminal navigation, we created a simulated environment that serves as a testbed for training and evaluating the Q-learning agent. 

\subsection{Simulation Environment}

\begin{figure}[ht]
\begin{center}
\includegraphics[width=1\columnwidth]{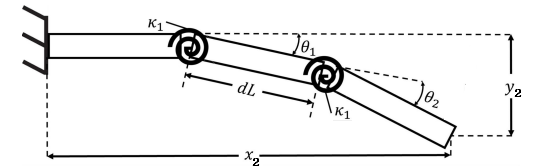}
\caption{The endoluminal robot's kinematic model. }
\label{fig:model}
\end{center}
\end{figure}

The bronchial tubes are assumed to be two-dimensional and are represented as a series of connected segments, each with specific properties such as curvature and diameter. The endoluminal robot can bend and move within the bronchial tubes while navigating toward predefined goals. 
We employ a model resembling an open-chain serial robot consisting of $N$ connected links, each featuring a rotational spring at its joint with a spring constant $k_1$ (see Figure \ref{fig:model}). The robot therefore undergoes dynamic motion until it eventually gains a minimal energy configuration.

To establish a proper robot behavior consider the configuration space $\C$. 
A configuration $c\in\C$ of the robot is the ordered set of its bending angles $c=\{\theta_i\}_{i=1}^N$. Accordingly, the positions $(x_i,y_i)_{i=1}^N$ of each joint may be calculated as depicted in Figure \ref{fig:model}. 

The energy of a free endoluminal robot is the sum of its spring energies $k_1\theta_i^2/2$. When the robot applies force on the lumen in $M$ positions, and therefore deforms the lumen in these positions, additional energy is added:
\begin{equation}
    E=\frac{1}{2}\sum_{i=1}^N k_1\theta_i^2+\frac{1}{2}\sum_{j=1}^M k_2(\Delta x_j^2+\Delta y_j^2) 
\end{equation}
where $\Delta x_j(\{\theta_i\}_{i=1}^j)$ and $\Delta y_j(\{\theta_i\}_{i=1}^j)$ is the extent at which the $j$-th contact exceeds the original position of the lumen (i.e. deforms it). The associated lumen spring constant is considered constant $k_2$ along the entire lumen.  

In every time step, the robot is provided with a relaxation stage at which it regains its minimal energy value. In other words, the robot is set to follow its energy gradient:
\begin{equation}
    \nabla E=(\frac{\partial E}{\partial\theta_1},\frac{\partial E}{\partial\theta_2},\dots,\frac{\partial E}{\partial\theta_N}) 
\end{equation}
and so in the relaxation stage, one updates the configuration $c'\leftarrow c+ \varepsilon\nabla E$ until an energy plateau is reached. Here, $\varepsilon$ is chosen small enough to account how interact is the configuration in the lumen. 

This procedure is then followed by bending of the robot tip and advancing (by holding its proximal end) into the lumen.

The agent's inability to achieve some goals prompted us to investigate the energy gradient used to calculate bending angles. By reducing the frequency of energy gradient adjustments, we aimed to improve the agent's ability to navigate and bend within the bronchial tubes. Subsequent trials with the modified energy gradient showed improvements, emphasizing the importance of parameter tuning in achieving successful navigation.

\subsection{Q-learning entities}

In the simulation, the \emph{state space} is defined by the agent's position and orientation within the bronchial tubes. The agent's position is determined by the \((x, y)\) coordinates of its distal end, while its orientation is defined by its angle of bending. This results in a six-dimensional state space.

The \emph{action space} consists of possible movements that the agent can perform, such as bending and advancing. The agent can bend in either the clockwise (CW) or counterclockwise (CCW) direction, and it can also advance forward. The agent's bending angle is limited by the physical constraints of the bronchial tubes.

The \emph{set goal} was to navigate to specific locations within the tubes. These goals were set as states within the environment, and the agent's objective was to learn how to reach these a predefined goal effectively.

\begin{figure}[t]
\begin{center}
\includegraphics[width=0.8\columnwidth]{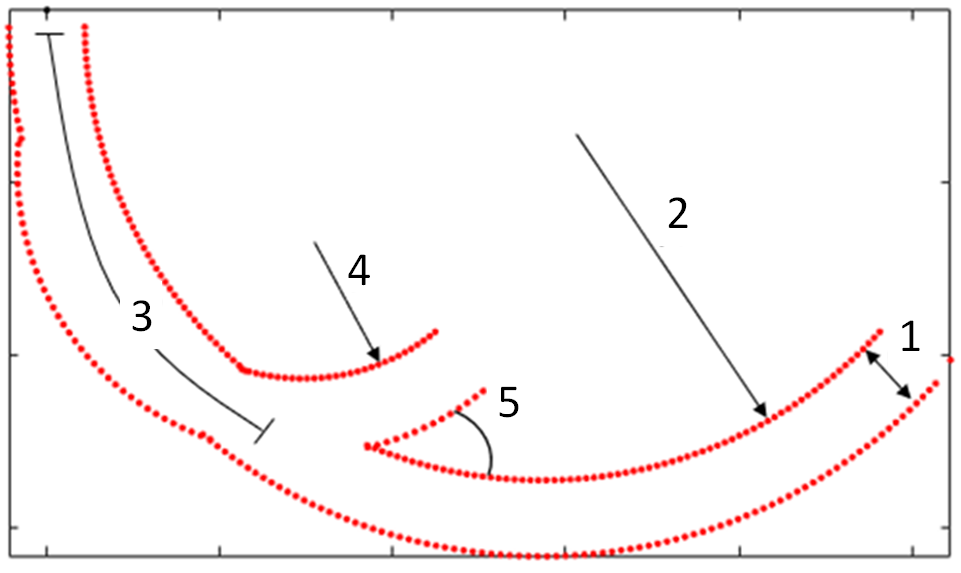}
\caption{The five parameters used to define a bifurcation: (1) the lumens' diameters; (2) the radius of curvature of the main lumen (can be positive or negative);(3) the distance to the bifurcation; (4) the radius of curvature of the bifurcating lumen (can be positive or negative); and (5) the bifurcation angle.}
\label{fig:bifurcation}
\end{center}
\end{figure}

\section{Methods and Results}
We conducted a set of $50$ experiments in four different lumen setup generated by a set of five randomly generated parameters (see Figure \ref{fig:bifurcation}) which define a lumen with a single bifurcation in the plane. The parameters define the lumen's diameters; the radius of curvature of the main lumen (can be positive or negative); the distance to the bifurcation; the radius of curvature of the bifurcating lumen (can be positive or negative); and the bifurcation angle.
The results of the Q-learning simulation indicated that the agent was able to learn effective navigation strategies within the bronchial tubes environment. The agent's success rate in reaching the predefined location was above $70\%$ after completing the learning phase.

 \begin{figure}[ht]
\begin{center}
\includegraphics[width=1\columnwidth]{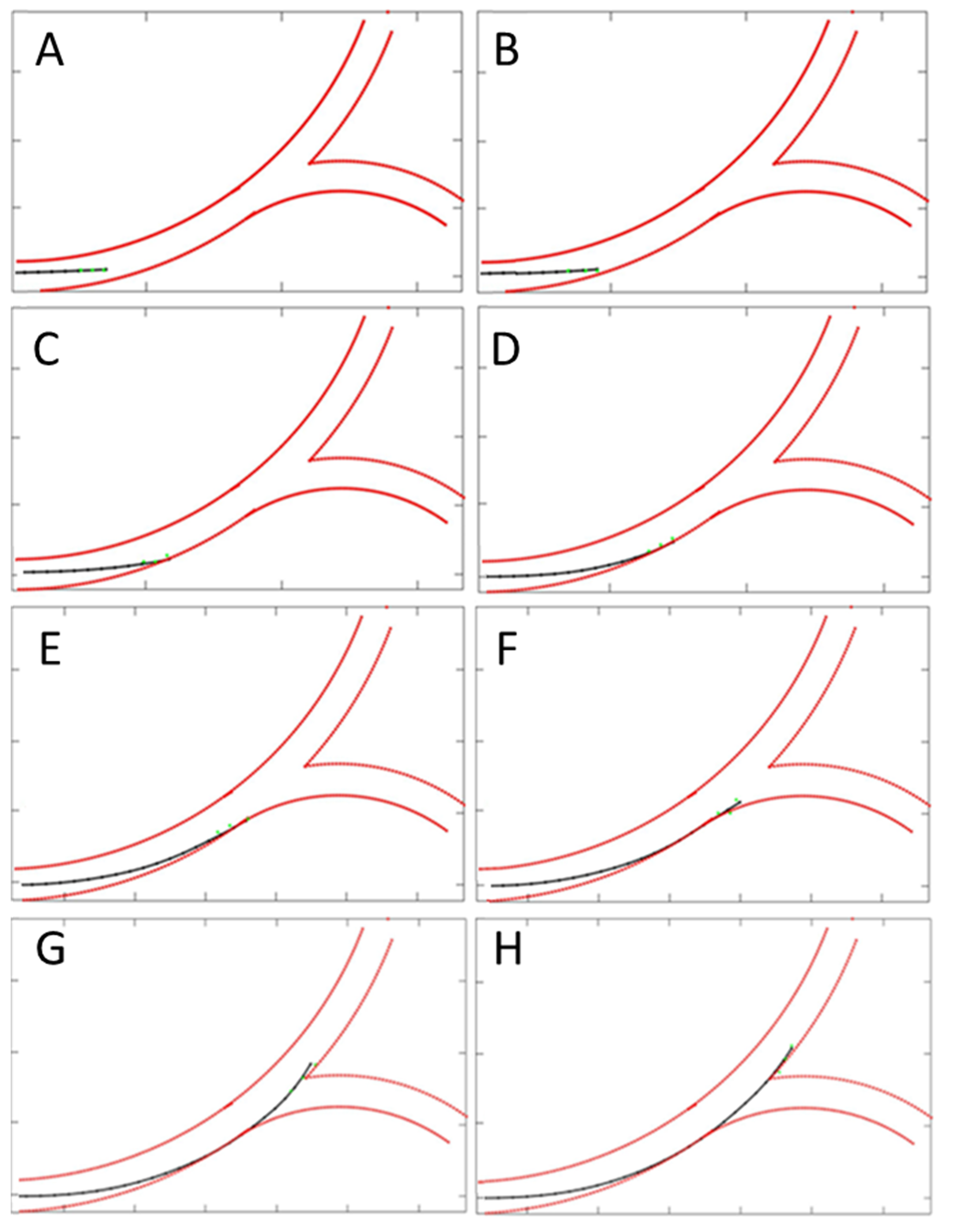}
\caption{A set of snapshots while maneuvering to the set goal through a given environment.}
\label{fig:3}
\end{center}
\end{figure}

 \begin{figure}[ht]
\begin{center}
\includegraphics[width=1\columnwidth]{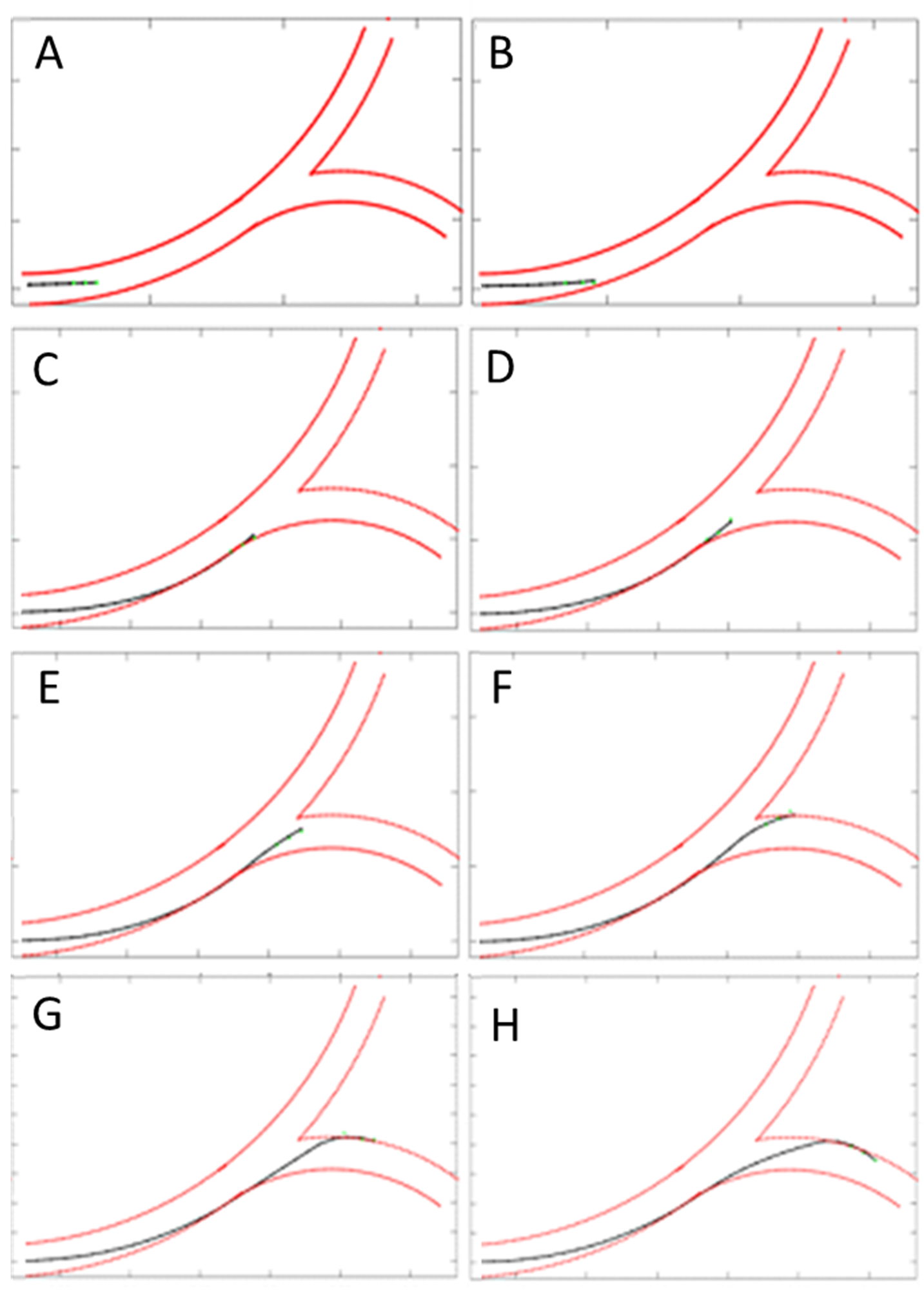}
\caption{A set of snapshots while maneuvering to the set goal.}
\label{fig:5}
\end{center}
\end{figure}

\section{Discussion}
The robotic agent was trained to choose appropriate actions to reach the predefined goals by estimating the Q-values for state-action pairs.  While the environment was discredited to facilitate learning, the agent still required thousands of sessions to effectively navigate. 

After the learning phase, the agent applied the epsilon greedy algorithm to balance exploration and exploitation, preventing it from getting stuck in infinite loops between states. 
 
The results indicated that the Q-learning agent could successfully learn navigation strategies in a complex environment.

The results of our experiments have broader implications for the application of Q-learning in medical procedures and navigations. While the current study focused on endoluminal navigation, the insights gained can be extended to various medical scenarios that require navigation within complex and constrained environments. 

\section{Conclusions}
The two-dimensional simulation environment deviates from the real-life conditions of navigating through three-dimensional bronchial tubes. As such, the results obtained from the simulation may not directly translate to real-world scenarios. Additionally, various parameters, such as the definition of terminal states and rewards, as well as the energy parameters, can impact the agent's behavior and performance. Further research is needed to explore the effects of these parameters and validate the results in more realistic settings.

The next phase of our project entails a more rigorous validation of our approach through simulations involving real-life scenarios, such as the replication of lung, kidney, and gall bladder environments. These simulations will be constructed using CT scans as our foundational data. Achieving success in this phase will necessitate the adaptation of our algorithm and the training of our agent to maneuver within physical bronchial tubes. This endeavor introduces an additional layer of complexity as it calls for the incorporation of a 3D model, granting the system an extra degree of freedom. This enhanced model will need to account for various factors, including friction, forces applied to the luminal walls, and the unpredictable conditions inherent to these biological environments.

Investigating the feasibility of transfer learning to adapt navigation strategies learned in simulated environments to real-world scenarios could expedite the deployment of Q-learning-based navigation systems.
\bibliographystyle{unsrt}
\bibliography{bib}








\end{document}